\documentclass[10pt,twocolumn,letterpaper]{article}

\usepackage[]{cvpr}      % To produce the REVIEW version
\usepackage{graphicx}
\usepackage{amsmath}
\usepackage{amssymb}
\usepackage{booktabs}
\usepackage{enumerate}
\graphicspath{{figures/}}
\usepackage{multirow}
\usepackage{booktabs}
\usepackage{bm}
\usepackage[switch]{lineno}
\usepackage{url}
\usepackage{makecell}
\usepackage{color}
\usepackage{pifont}
\usepackage[pagebackref,breaklinks,colorlinks]{hyperref}

\usepackage[capitalize]{cleveref}
\crefname{section}{Sec.}{Secs.}
\Crefname{section}{Section}{Sections}
\Crefname{table}{Table}{Tables}
\crefname{table}{Tab.}{Tabs.}

\begin{document}

\title{Stereo Image Rain Removal via Dual-View Mutual Attention}

\author{Yanyan Wei, Zhao Zhang*, Zhongqiu Zhao, Yang Zhao and Richang Hong \vspace{1.5mm}\\ 
	Hefei University of Technology, Hefei, China\\
	{\tt\small E-mails: weiyanyan@mail.hfut.edu.cn, cszzhang@gmail.com}
\and
Yi Yang\\
Zhejiang University, Hangzhou, China\\
}
%\maketitle

\twocolumn[{
	\renewcommand\twocolumn[1][]{#1}
	\maketitle
	\begin{center}
		\captionsetup{type=figure}
		\vspace{-8mm}
		\begin{minipage}{1\linewidth}
			\centering
			\includegraphics[width=1\linewidth,height=0.1\linewidth]{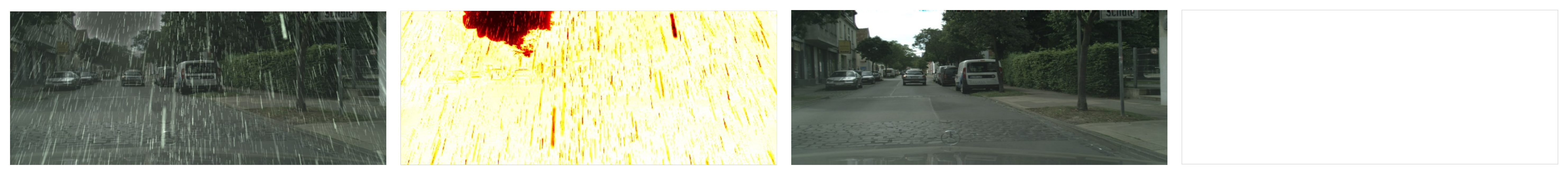}
		\end{minipage}\\
%		\vspace{-2mm}
		\begin{minipage}{1\linewidth}
			\centering\vspace{1pt}
			\begin{minipage}{0.245\linewidth} \centering \footnotesize (a) Rain image \end{minipage}
			\begin{minipage}{0.245\linewidth} \centering \footnotesize (b) Error map of rain image \end{minipage}
			\begin{minipage}{0.245\linewidth} \centering \footnotesize (c) Ground truth \end{minipage}
			\begin{minipage}{0.245\linewidth} \centering \footnotesize (d) Error map of ground truth \end{minipage}
		\end{minipage}
		\begin{minipage}{1\linewidth}
			\centering
			\includegraphics[width=1\linewidth,height=0.1\linewidth]{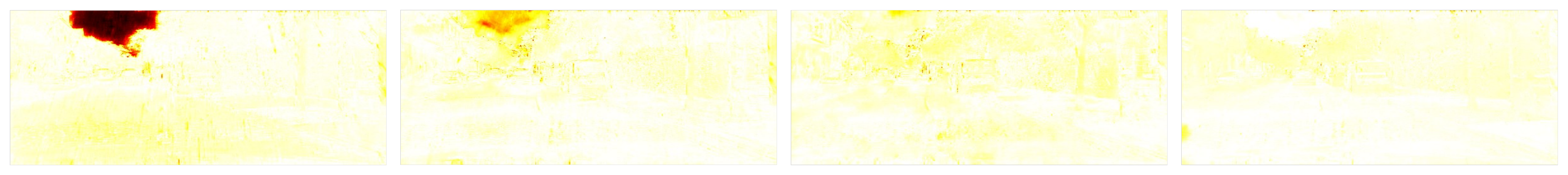}
		\end{minipage}\\	
		\begin{minipage}{1\linewidth}
			\centering\vspace{1pt}
			\begin{minipage}{0.245\linewidth} \centering \footnotesize (e) RESCAN \end{minipage}
			\begin{minipage}{0.245\linewidth} \centering \footnotesize (f) JORDER-E \end{minipage}
			\begin{minipage}{0.245\linewidth} \centering \footnotesize (g) RCDNet \end{minipage}
			\begin{minipage}{0.245\linewidth} \centering \footnotesize (h) MPRNet \end{minipage}
		\end{minipage}
		\begin{minipage}{1\linewidth}
			\centering
			\includegraphics[width=1\linewidth,height=0.1\linewidth]{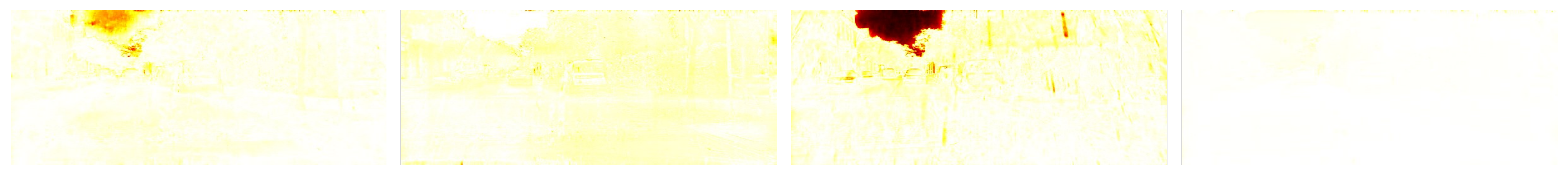}
		\end{minipage}\\	
		\begin{minipage}{1\linewidth}
			\centering\vspace{1pt}
			\begin{minipage}{0.245\linewidth} \centering \footnotesize (i) iPASSR \end{minipage}
			\begin{minipage}{0.245\linewidth} \centering \footnotesize (j) NAFSSR \end{minipage}
			\begin{minipage}{0.245\linewidth} \centering \footnotesize (k) EPRRNet \end{minipage}
			\begin{minipage}{0.245\linewidth} \centering \footnotesize (l) StereoIRR \end{minipage}
		\end{minipage}
		\vspace{-4mm}
		\captionof{figure}{Visualization comparison of the produced error maps by each method on the StereoCityscapes dataset, including both monocular methods (i.e., RESCAN \cite{RESCAN}, JORDER-E \cite{JORDER-E}, RCDNet \cite{RCDNet} and MPRNet \cite{MPRNet}) and stereo methods (i.e., iPASSR \cite{iPASSR}, NAFSSR \cite{NAFSSR}, EPRRNet \cite{EPRRNet} and our StereoIRR). The error map is obtained by calculating the difference between the derained image of each method and the ground-truth image \cite{Error}. For the error maps, the whiter the better, e.g., the rain image has the darkest error map, while the ground-truth image has the whitest error map. Clearly, our StereoIRR is capable of preserving the structure and recovering the details more accurately, which can also be concluded from the produced smaller error.}
		\label{fig:1}
		\vspace{2mm}
	\end{center}
}]

\begin{abstract}
\vspace{-4mm}
Stereo images, containing left and right view images with disparity, are utilized in solving low-vision tasks recently, e.g., rain removal and super-resolution. Stereo image restoration methods usually obtain better performance than monocular methods by learning the disparity between dual views either implicitly or explicitly. However, existing stereo rain removal methods still cannot make full use of the complementary information between two views, and we find it is because: 1) the rain streaks have more complex distributions in directions and densities, which severely damage the complementary information and pose greater challenges; 2) the disparity estimation is not accurate enough due to the imperfect fusion mechanism for the features between two views. To overcome such limitations, we propose a new \underline{Stereo} \underline{I}mage \underline{R}ain \underline{R}emoval method (StereoIRR) via sufficient interaction between two views, which incorporates: 1) a new Dual-view Mutual Attention (DMA) mechanism which generates mutual attention maps by taking left and right views as key information for each other to facilitate cross-view feature fusion; 2) a long-range and cross-view interaction, which is constructed with basic blocks and dual-view mutual attention, can alleviate the adverse effect of rain on complementary information to help the features of stereo images to get long-range and cross-view interaction and fusion. Notably, StereoIRR outperforms other related monocular and stereo image rain removal methods on several datasets. Our codes and datasets will be released. 
\end{abstract}

\vspace{-4mm}%不可以换行
\section{Introduction}
With the development of hardware equipment, stereo image data is no longer scarce and promote the development of stereo restoration research quickly. Stereo image restoration methods \cite{Stereo_DB1,Stereo_DH1,EPRRNet,NAFSSR,iPASSR} mainly focus on using the complementary information between the left and right view images to recover accurate structure and texture information from the degraded images, which benefits the visual perception and subsequent high-level understanding tasks. For example, Zhang \textit{et al.} \cite{PRRNet} presented a paired rain removal network named PRRNet, which exploits both stereo images and semantic information and fuses them with a parallel network. To the best of our knowledge, this is the first attempt to use deep learning to remove rain streaks from stereo images. After that, Shi \textit{et al.} \cite{Stereo_DR5} focused on detecting and removing raindrops from stereo images, and proposed a row-wise dilated attention module to enlarge the attention’s receptive field for effective information propagation between different views. 

Despite some stereo image restoration methods \cite{PRRNet,EPRRNet,Stereo_DR5,iPASSR} stated that stereo methods are promising for the rain removal task, their performances are still inferior to some monocular rain removal methods \cite{MPRNet,RCDNet}. It is obviously indicated in Figure \ref{fig:1} that these stereo methods even bring larger errors than those monocular methods, meaning they are not capable of preserving the structure and recovering the details more accurately. We hypothesize that such performance gap roots in two main challenges of stereo rain removal task: 1) unlike stereo super-resolution methods \cite{Stereo_SR1,Stereo_SR4,Stereo_SR6,NAFSSR,iPASSR}, which can use accurate dual-view image information, stereo rain removal methods \cite{PRRNet,EPRRNet} face greater difficulty, since complex rain streaks with a high diversity of rain directions and types will greatly damage the image details of both left and right view; 2) the horizontal disparity under dual-constraints and the complexity of the rain streak itself will bring great difficulty to the information interaction and feature fusion, and unreliable disparity estimation results in the loss of detail and structure of derained images. Overall, the complex distribution of rain streaks will bring great challenges to disparity estimation and feature fusion for stereo image rain removal task, and even cause the performance worse than the monocular image rain removal methods \cite{MPRNet,RCDNet,JORDER-E}. 

\begin{figure}[t!]% 处引用 
	\setlength{\unitlength}{1cm}
	\begin{center}
		\centering
		\includegraphics[width=1\linewidth]{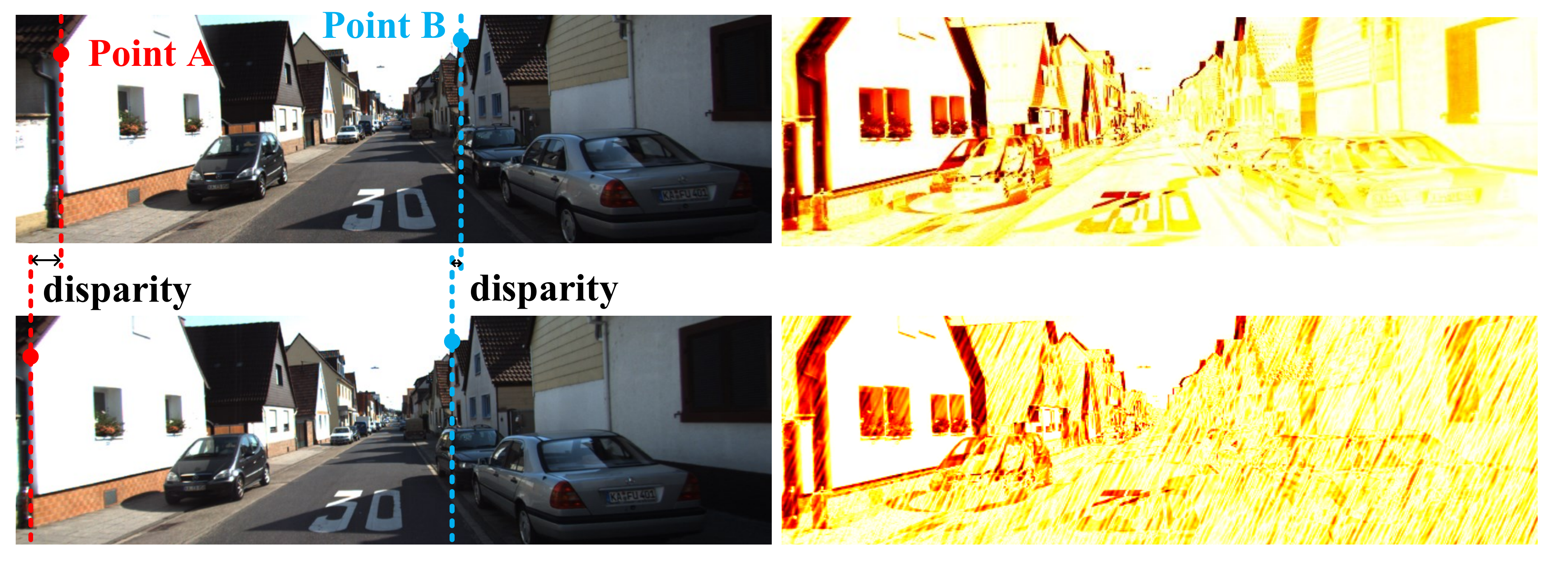}	
		\begin{picture}(0,0)
			\put(-2.1,0.2){(a)}
			\put(1.9,0.2){(b)}
%			\put(-1.7,0.2){(c) epipolar constraint}
%			\put(-3.2,2.2){\tiny $\mathbf{X}^{l}$(\textit{h},\textit{w})}
%			\put(0.6,2.2){\tiny $\mathbf{X}^{r}$(\textit{h},\textit{w+d})}
		\end{picture}    
	\end{center}
	\vspace{-6.5mm}%不可以换行
	\caption{Two characteristics of the stereo image rain removal task. (a) shows the left-view and right-view images from the RainKITTI2012 dataset respectively, from which we can see that the disparity of point A which is closer to the lens (i.e., the scene depth is lower) is much larger than that of point B (i.e., the disparity is inversely proportional to the scene depth). (b) shows the error maps by calculating the difference between left and right view images. The first row of (b) is obtained with clean stereo images, while the second row of (b) is obtained with rain stereo images, from which we can see that rain affects the expression of complementary structural information.}
	\label{fig:2}
	\vspace{-4mm}%不可以换行
\end{figure}
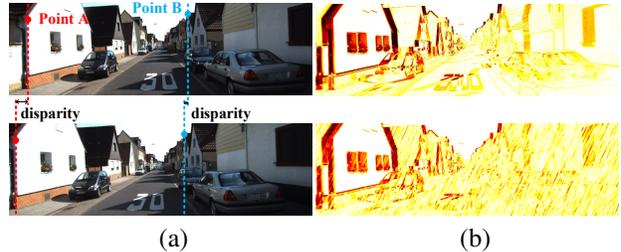 

To verify our hypotheses, we investigate the stereo image rain removal task and found the following characteristics: 1) the horizontal disparity of left and right view images under epipolar constraint is inversely proportional to the scene depth, i.e., the closer the position is to the lens, the larger the disparity will be; 2) usually, we can obtain the structural information of the scene based on the complementary relationship between the left and right views, but the rain streaks will obviously destroy this structural information. Figure \ref{fig:2} shows our investigations clearly, from which we can derive the following conclusions. Firstly, since the disparity between left and right views is related to the scene depth and is not fixed, model must have the ability to extract features with different scales in order to perform better disparity estimation; secondly, rain streaks have obvious influence on the complementary information expression, so model need have a long-range process to alleviate the effects of rain streaks; thirdly, an efficient mechanism needs to be designed to take advantage of the complementary information from both views, in order to perform better cross-view fusion. We are then motivated to design a new stereo image rain removal method based on the above investigations. A \textbf{L}ong-range and \textbf{C}ross-view \textbf{I}nteraction (\textbf{LCI}) process is presented to help the features of stereo images to get long-range and cross-view interaction fusion. Specifically, LCI uses a U-net \cite{U-net} architecture as the backbone, which is constructed with BasicBlocks and Dual-view Mutual Attention (DMA). BasicBlocks extract features from two view images using different scales and shortcuts, while DMA calculates dual-view mutual attention maps and conducts cross-view fusion by having left and right views as \textit{key} matrix for each other. This setting enables our method to preserve the structure and recover the details more accurately, as can be seen in Figure \ref{fig:1}. 

Overall, the contributions are summarized as follows: 

\begin{enumerate}
	\vspace{-2mm}%不可以换行
	\item \textbf{StereoIRR: A New and Effective Stereo Image Rain Removal Method}. We propose a novel stereo image rain removal strategy based on a novel dual-view mutual attention mechanism. Specifically, a long-range and cross-view interaction process is constructed with BasicBlocks and dual-view mutual attention, which can alleviate the adverse effect of rain on complementary information to help the features of stereo images to get long-range and cross-view interaction and fusion. 
	\vspace{-2mm}%不可以换行
	\item \textbf{Dual-view Mutual Attention Mechanism}. We propose a dual-view mutual attention mechanism that utilizes feature refinement from both channel and spatial perspectives to learn the disparity map of stereo view implicitly. Unlike single-view self attention, our dual-view mutual attention can generate mutual attention maps by having left and right views as the \textit{key} matrix for each other to facilitate cross-view feature fusion. 
	\vspace{-2mm}%不可以换行
	\item \textbf{SOTA Performance on Rain Removal}. StereoIRR outperforms other related monocular and stereo image rain removal methods. StereoIRR achieves more than 3 dB PSNR improvement over stereo method EPRRNet \cite{EPRRNet} on the RainKITTI2012 and RainKITTI2015 datasets, and is also superior to the SOTA monocular method MPRNet \cite{MPRNet} with 1.5 dB PSNR improvement on three widely-used image datasets.
	\vspace{-4mm}%不可以换行
\end{enumerate}

\begin{figure*}[ht!]	
	\setlength{\unitlength}{1cm}
	\begin{center}
		\centering
		\includegraphics[width=0.9\linewidth]{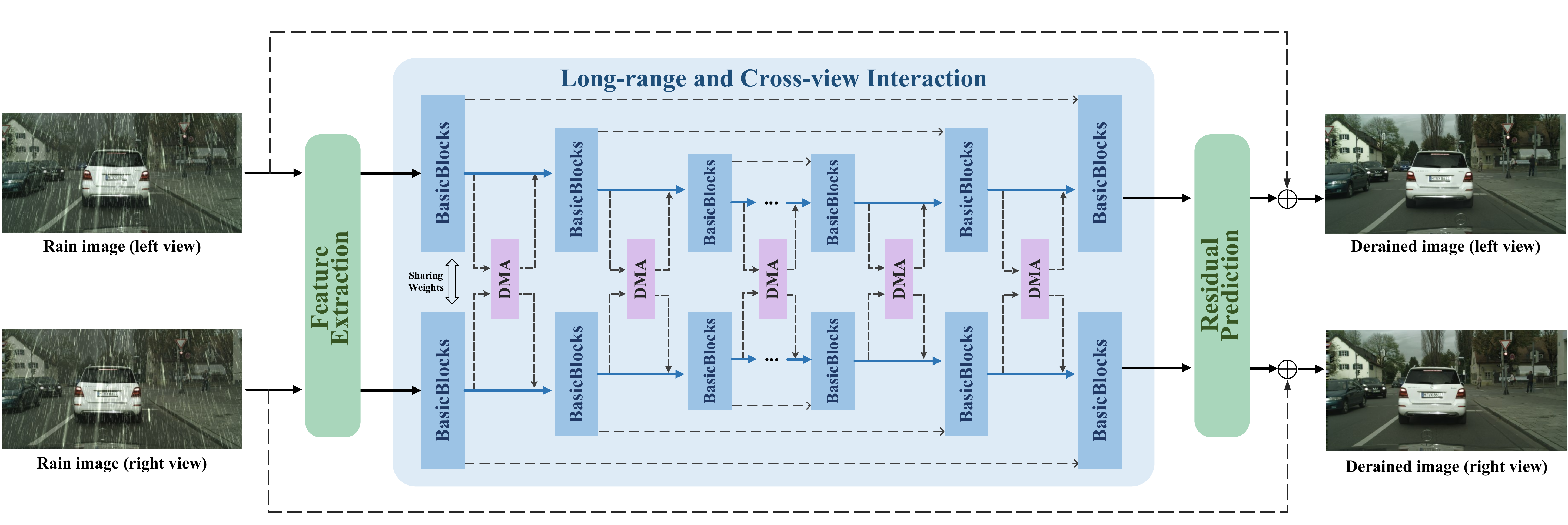}	
		\begin{picture}(0,0)
		\end{picture}    
	\end{center}
	\vspace{-8mm}%不可以换行
	\caption{The pipeline of our StereoIRR consists of three processes: feature extraction, long-range and cross-view interaction (LCI), and residual prediction. LCI is constructed with BasicBlocks and dual-view mutual attention (DMA), which can help the features of stereo images to get long-range and cross-view interaction and fusion.}
	\label{fig:3}
	\vspace{-4mm}%不可以换行
\end{figure*} 

\section{Related Work}
\subsection{Monocular Image Deraining}
Traditional model/prior-based monocular image deraining (MID) methods \cite{DSC,GMM} fail to accurately separate the rain layer and background layer, which has been gradually surpassed by the deep learning-based methods \cite{SGINet,UBCN,RadNet,CCN,DerainCycleGAN,SIRR,Uformer,Restormer,DetailNet,RCDNet,RESCAN,JORDER-E,MPRNet} in recent years. The main reasons for the performance improvement are twofold: 1) deep neural networks have powerful feature extraction and mapping capabilities; 2) a large amount of training data provide sufficient information for network training. To be specific, MID methods mainly consider removing the rain streak component $R$ and recovering the clean background image $B$ from a rain image $X$, which is formulated as $X=R+B$.

End-to-end deep MID methods use different deep neural networks as backbones, such as convolutional neural network \cite{VGG,U-net,CA}, recurrent neural network \cite{LSTM} and adversarial generative network \cite{GAN}, to extract the hierarchical features and deep information of the rain images, and get a direct mapping from rain image to clear image. Different basic blocks or modules are designed for deep feature extraction, such as residual or dense block \cite{RESCAN,PRRNet,DeHRain}, hybrid block \cite{RadNet}, channel attention module \cite{MSPFN}, spatial attentive module \cite{SPANet}, self-attention \cite{Restormer} and transformer \cite{Uformer,SwinIR}, etc. To train a better deep model, prior knowledge and side information related to the rain or background layer is also utilized, such as rain mask \cite{JORDER-E}, rain density \cite{DID-MDN}, rain speed \cite{RICNet}, rain kernel \cite{RCDNet}, semantic segmentation \cite{UBCN,SGINet}, scene depth \cite{DAF-Net}, transmission map \cite{DeHRain}, and so on. 

\subsection{Stereo Image Deraining and Restoration}
Unlike the MID methods that use a single image as input, Stereo image deraining (SID) methods aim to process both left and right views with disparity. In theory, stereo images can provide more detail and structural information than monocular images. However, the main difficulty lies in appropriately fusing the features of left and right view images with disparity and ensuring consistency in stereo images. So far, only rare prior works were done, i.e., \cite{PRRNet,EPRRNet,Stereo_DR5}. Shi \textit{et al.} \cite{Stereo_DR5} proposed a novel row-wise dilated attention module to enlarge attention’s receptive field for effective information propagation between the two stereo images. Zhang \textit{et al.} \cite{PRRNet} present a stereo image restoration method PRRNet and its enhanced version EPRRNet \cite{EPRRNet} to exploit both stereo images and semantic information. EPRRNet contains a semantic-aware deraining module that solves both tasks of semantic segmentation and deraining of scenes, a semantic-fusion network that fuses semantic information and derained information, and a view-fusion network that fuses multi-view information, respectively. Note that existing methods \cite{PRRNet,EPRRNet,Stereo_DR5} use an implicit way to learn the disparity map of stereo view. 

In addition to the rain removal task, stereo images also benefit other tasks, e.g., image super-resolution \cite{Stereo_SR1,Stereo_SR4,Stereo_SR6}, image dehazing \cite{Stereo_DH1,Stereo_DH2} and image deblurring \cite{Stereo_DB1,Stereo_SR4}. Jeon \textit{et al.} \cite{Stereo_SR1} adopted the translation copy stacking method to realize the fusion of complementary information between left and right views under different disparities, but this artificially set disparity is not flexible and accurate. Some stereo super-resolution methods \cite{Stereo_SR4,Stereo_SR6} are based on disparity estimation subtask, i.e., firstly calculate the disparity of each spatial position in the left and right views, and then perform pixel-level registration of the left and right view image under the guidance of the disparity map. However, taking the disparity estimation as a subtask brings additional processing difficulty and computational cost. Some other stereo super-resolution methods \cite{iPASSR,NAFSSR} mine the prior constraints of symmetry between stereo images and implicitly learn the disparity using the attention mechanism. Based on the strong fitting ability of neural networks and the rationality of modules, these methods get better results.

\section{Proposed Method}
The architecture of our StereoIRR is shown in Figure \ref{fig:3}. In addition to the prepositive feature extraction and postpositional residual prediction processes, \textbf{L}ong-range and \textbf{C}ross-view \textbf{I}nteraction (\textbf{LCI}) is the most critical process of StereoIRR. LCI is constructed with BasicBlocks and Dual-view Mutual Attention (DMA), and can help the features of stereo images to get long-range and cross-view interaction and fusion. We will introduce each process below.

\subsection{Network Pipeline}
\textbf{Feature Extraction.}
Using shallow convolutional layers to quickly map images with RGB channels to a wider channels space can help subsequent modules learn better image features, which has been proved in previous works \cite{SwinIR,NAFNet,NAFSSR}. In StereoIRR, given a pair of stereo input $(X_l, X_r)\in\mathbb{R}^{H_{in} \times W_{in} \times C_{in}}$, we use a $3\times3$ convolutional layer $H_F\left ( \cdot \right )$ to extract shallow features $F_{l,r}^0\in\mathbb{R}^{2H_{in} \times 2W_{in} \times C}$ as 
\begin{equation} \label{eqn1}
	F_{l,r}^0 = H_F(X_l, X_r),
\end{equation}
\noindent where $X_l$ and $X_r$ denote the left and right view rain images. $F_{l,r}^0$ is the concatenate features of left-view features $F_l^0$ and right-view features $F_r^0$. $H_{in}$, $W_{in}$, and $C_{in}$ are the height, width, and channel number of the input image. $C$ is the basic feature channel number. In StereoIRR, $H_{in}=W_{in}=320$, $C_{in}=3$, and we set $C=30$. After obtaining the shallow features $F_{l,r}^0$ of both views, we feed them into the next process for long-range and cross-view interaction.

\textbf{Long-range and Cross-view Interaction (LCI).}
Unlike the existing monocultural image restoration methods \cite{iPASSR,NAFSSR}, which use pixel-invariant network structures, our StereoIRR uses a U-net \cite{U-net} structure with shortcuts as the backbone. We performed five up-down sampling operations in LCI for a long-range and cross-view interaction between stereo-view features. Specifically, the deep features of left and right views are extracted by BasicBlocks with shared parameters, and are then fused based on DMA. The details of BasicBlocks and DMA will be introduced later. The brief process of LCI can be described as follows:
\begin{equation} \label{eqn2}
	F_{l,r}^d = H_L(F_{l,r}^0),
\end{equation}
\noindent where $F_{l,r}^0$ are the shallow features extracted by prepositive feature extraction. $H_L\left ( \cdot \right )$ denotes the process of LCI. $F_{l,r}^d$ is the deep feature after long-range and cross-view interaction. 

\textbf{Residual Prediction.}
Resembling the feature extraction process, we also use a shallow convolutional layer to remap deep features to RGB channels. We predict the rain streak residuals and add them to the original stereo inputs to get more accurate stereo rain removal results: 
\begin{equation} \label{eqn3}
	\hat{Y}_l, \hat{Y}_r = H_R(F_{l,r}^d),
\end{equation}
\noindent where $H_R\left ( \cdot \right )$ denotes the process of residual prediction. $\hat{Y}_l$ and $\hat{Y}_r$ are the left-view and right-view rain removal images, respectively. 

\subsection{BasicBlocks}\label{BasicBlocks}
\begin{figure}[t!]% 处引用 
	\vspace{-4mm}%不可以换行
	\begin{center}
		\centering
		\includegraphics[width=\linewidth]{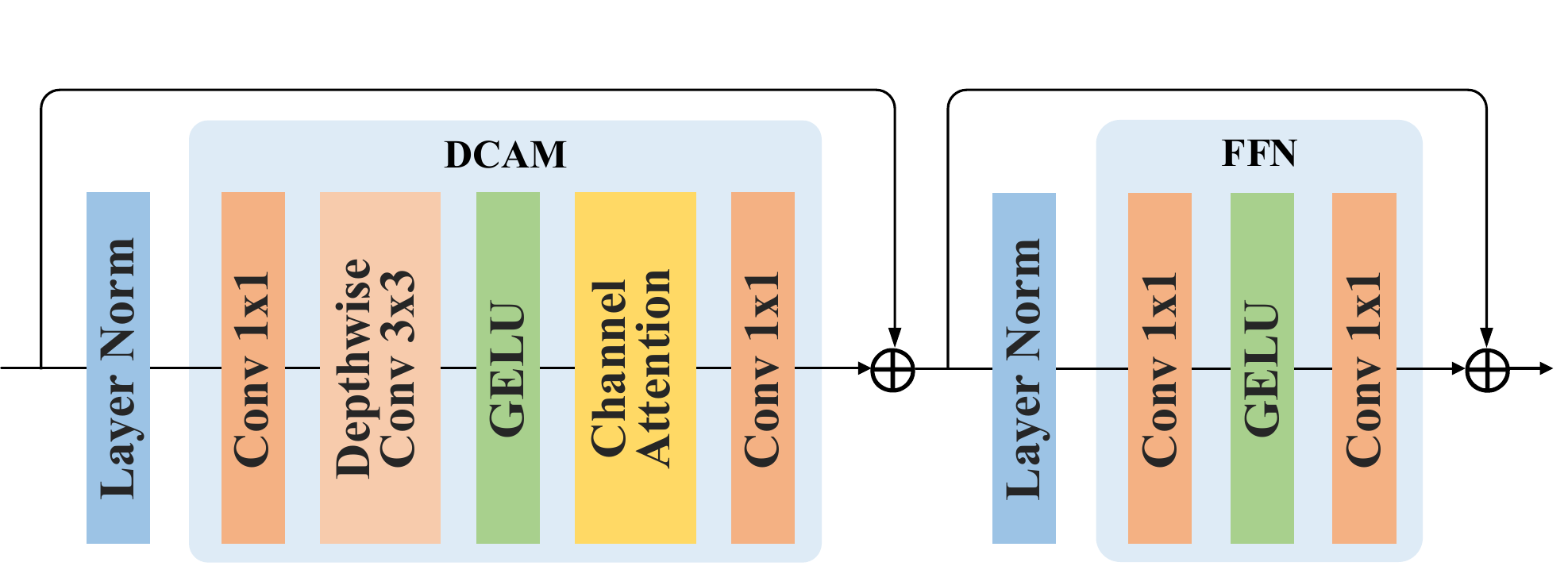}	
		\begin{picture}(0,0)	
		\end{picture}    
	\end{center}
	\vspace{-10mm}%不可以换行
	\caption{\mdseries \textbf{BasicBlocks} consists of two parts: Depth-wise Channel Attention Module (DCAM) and Feed-Forward Network (FFN).}
	\vspace{-4mm}%不可以换行
	\label{fig:4}
\end{figure}

The BasicBlocks is introduced by NAFNet \cite{NAFNet}, and its structure is illustrated in Figure \ref{fig:4}. There are two parts in BasicBlocks, i.e., Depth-wise Channel Attention Module (DCAM) and Feed-Forward Network (FFN). DCAM mainly consists of point-wise and depth-wise convolution layers with channel attention (CA) \cite{CA}. FFN has two point-wise convolution layers and a nonlinear layer (GELU) \cite{GELU} added in the middle. The LayerNorm (LN) \cite{LN} layer is added before both DCAM and FFN, and the nonlinear layer (GELU) is added in the middle. The residual connection is employed for both modules using two learnable weight parameters. This long-range and inter-view feature extracted process by cascaded BasicBlocks can be formulated as
\begin{equation} \label{eqn4}
	\begin{split}
		F_{l,r}&=\alpha \cdot \mathrm{DCAM}\left (\mathrm{LN}\left ( F_{l,r}^{i-1}\right ) \right )+F_{l,r}^{i-1},\\
		F_{l,r}^{i}&=\beta \cdot \mathrm{FFN}\left (\mathrm{LN}\left ( F_{l,r}\right ) \right )+F_{l,r},
	\end{split}
\end{equation}
\noindent where $F_{l,r}^{i-1}$ and $F_{l,r}^i$ are the stereo-view feature extracted from the prepositive and current BasicBlocks, respectively. $\alpha$ and $\beta$ are two trainable channel-wise scales and initialized with zeros for stabilizing training. We split the current stereo-view feature $F_{l,r}^i$ into $F_l^i$ and $F_r^i$ and send them into DMA for cross-view feature fusion.

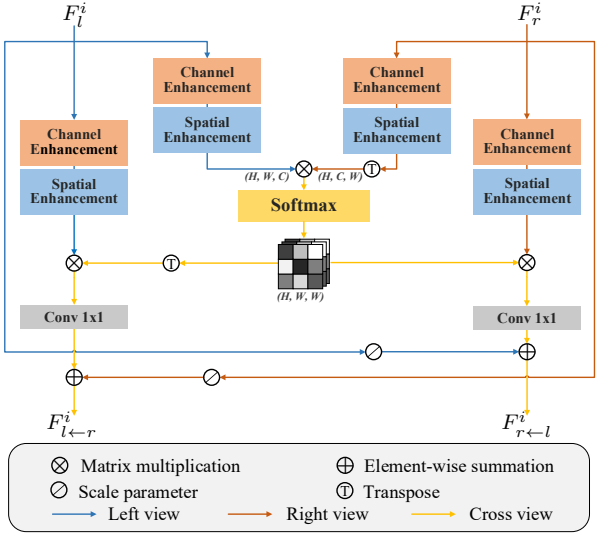
\begin{figure}[t!]% 处引用 
	\vspace{-2mm}%不可以换行
	\begin{center}
		\centering
		\includegraphics[width=\linewidth]{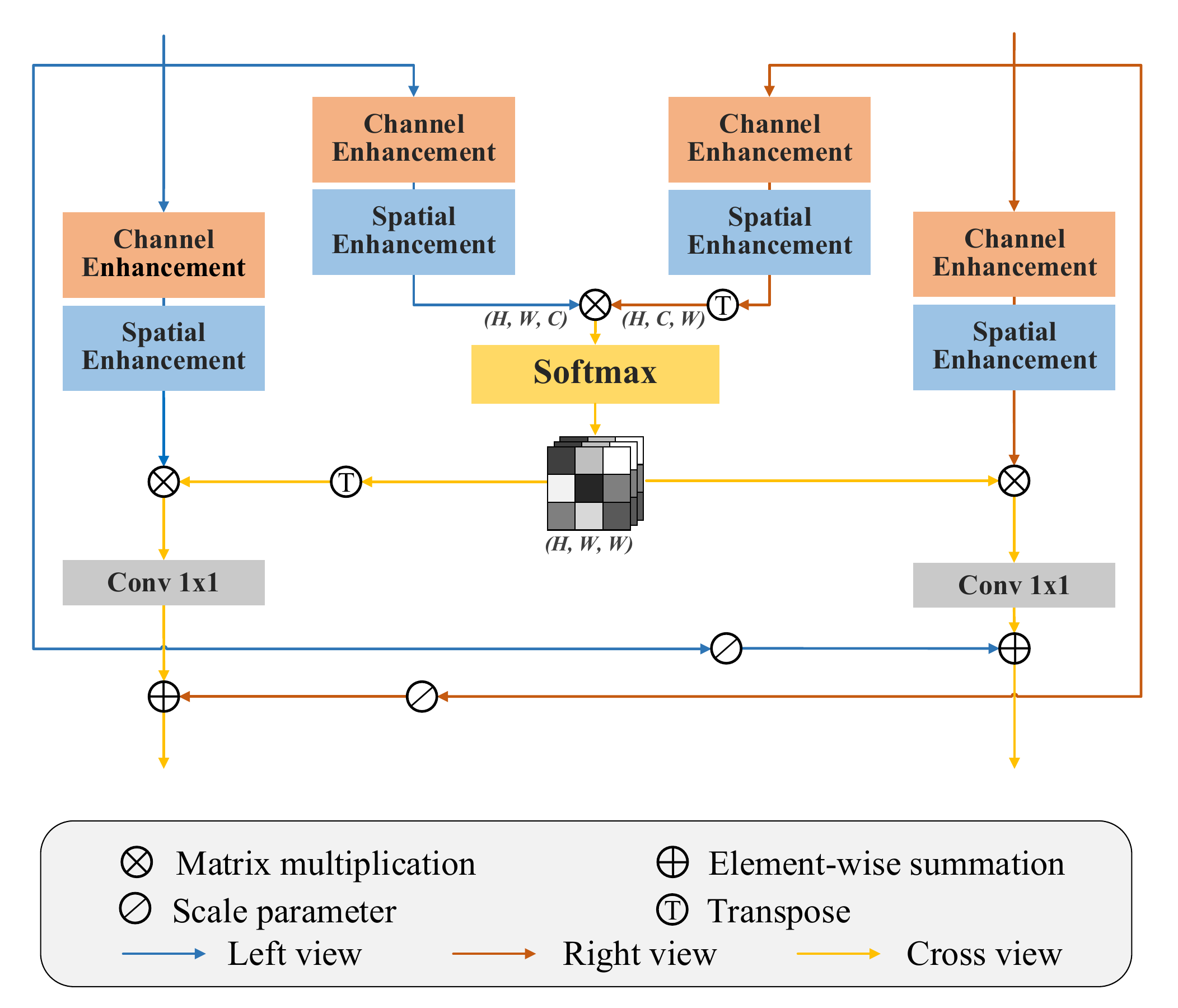}	
		\begin{picture}(0,0)
			\put(-90,210){\footnotesize $F_l^i$}
			\put(82,210){\footnotesize $F_r^i$}
			\put(-96,55){\footnotesize $F_{l\leftarrow r}^i$}
			\put(76,55){\footnotesize $F_{r\leftarrow l}^i$}
			%\put(-84,142){\tiny $\mathrm{Attention}_{l\leftarrow r}$}
			%\put(44,142){\tiny $\mathrm{Attention}_{r\leftarrow l}$}
		\end{picture}    
	\end{center}
	\vspace{-10mm}%不可以换行
	\caption{\textbf{Dual-view Mutual Attention} (DMA) can generate cross-view attention by having left and right views as \textit{key} matrix for each other. To obtain accurate feature projection, we conduct feature refinement via both channel and spatial enhancement.}
	\vspace{-4mm}%不可以换行
	\label{fig:5}
\end{figure} 

\subsection{Dual-view Mutual Attention (DMA)}\label{DMA}
Some previous image restoration works \cite{Stereo_SR4,Stereo_SR6} take disparity estimation as a subtask and use the estimated disparity to guide the feature fusion of left and right views. However, the unreliable disparity will have a negative impact on the restoration task, and disparity estimation will also bring additional computational costs. In order to avoid the above disadvantages, we design a DMA mechanism that uses dual-view features to learn disparity implicitly and fuses the stereo features extracted from the BasicBlocks. 

\textbf{Single-view self-attention.}
The self-attention mechanism has been widely used in previous works \cite{Attention,Restormer}, and its original process can be described as follows: 
\begin{equation} \label{eqn5}
	\mathrm{Attention}\left ( \mathbf{Q},\mathbf{K},\mathbf{V}\right )=\mathrm{softmax}\left ( \mathbf{Q} \mathbf{K}^T/\sqrt{d_k}\right ) \mathbf{V}, 
\end{equation}%
\noindent where $\mathbf{Q}$, $\mathbf{K}$ and $\mathbf{V}$ are the projected matrices \textit{query}, \textit{key}, and \textit{value}, respectively. $d_k$ is the dimension of queries and keys. Eqn. (\ref{eqn5}) computes the dot products of the \textit{query} with all \textit{keys}, divide each by $\sqrt{d_k}$, and employs the softmax function to obtain the weights on the \textit{value}.

\textbf{Dual-view mutual attention.}
Different from monocular restoration methods \cite{Restormer,SwinIR}, which use single-view features to generate self-attention maps, our StereoIRR can generate cross-view attention maps based on left and right-view images via a designed dual-view mutual attention mechanism. The proposed DMA is illustrated in Figure \ref{fig:5}. Due to the epipolar constraint \cite{iPASSR}, there is only horizontal disparity between the left and right views. We take advantage of the features of the stereo images by having left and right views as \textit{key} ($\mathbf{K}$) matrix for each other
\begin{equation} \label{eqn6}
	\mathrm{Attention}_{l\leftarrow r}\left ( \mathbf{Q_l},\mathbf{K_r},\mathbf{V_l}\right )=\mathrm{softmax}\left ( \mathbf{Q_l} \mathbf{K_r}^T/\sqrt{d_k}\right ) \mathbf{V_l},\\
\end{equation}
\noindent where the subscript (\textbf{l} or \textbf{r}) of $\mathbf{Q}$, $\mathbf{K}$ and $\mathbf{V}$ indicates which view they are projected from. Eqn. (\ref{eqn6}) computes the dot products of the left-view \textit{query} with right-view \textit{key}, divide each by $\sqrt{d_k}$, and applies a softmax function to obtain the weights on the left-view \textit{value}. Based on the calculated dual-view mutual attention $\mathrm{Attention}_{l\leftarrow r}$, we can get the cross-view fusion features $F_{l\leftarrow r}^i$ as follows: 
\begin{equation} \label{eqn7}
	\begin{split}
	F_{l\leftarrow r}^i=\gamma_1 W_p^l\mathrm{Attention}_{l\leftarrow r}\left ( \mathbf{Q_l},\mathbf{K_r},\mathbf{V_l}\right )+F_r^i, \\
	\mathbf{Q_l}=W_d^{Q_l}W_p^{Q_l}F_l^i, \mathbf{K_r}=W_d^{K_r}W_p^{K_r}F_r^i, \mathbf{V_l}=W_d^{V_l}W_p^{V_l}F_l^i,
	\end{split}
\end{equation}
\noindent where $\gamma_1$ is a trainable channel-wise scale parameter and initialized with zeros for stabilizing training, $W_p^l$ is the 1$\times$1 point-wise convolution. We offer feature refinement from both channel and spatial perspectives, i.e., $W_p^{\left ( \cdot \right )}$ is the 1$\times$1 point-wise convolution for channel enhancement, and $W_d^{\left ( \cdot \right )}$ is the 3$\times$3 depth-wise convolution for spatial enhancement. $F_l^i$ is left-view features extracted by the front BasicBlocks.

With analogous argument, we can obtain the cross-view fusion features $F_{r\leftarrow l}^i$ of the right view based on the dual-view mutual attention $\mathrm{Attention}_{r\leftarrow l}$ as follows: 
\begin{equation} \label{eqn8}
	\begin{split}
	\mathrm{Attention}_{r\leftarrow l}\left ( \mathbf{Q_r},\mathbf{K_l},\mathbf{V_r}\right )=\mathrm{softmax}\left ( \mathbf{Q_r}^T \mathbf{K_l}/\sqrt{d_k}\right ) \mathbf{V_r}, \\
	F_{r\leftarrow l}^i=\gamma_2 W_p^r\mathrm{Attention}_{r\leftarrow l}\left ( \mathbf{Q_r},\mathbf{K_l},\mathbf{V_r}\right )+F_l^i, \\
	\mathbf{Q_r}=W_d^{Q_r}W_p^{Q_r}F_r^i, \mathbf{K_l}=W_d^{K_l}W_p^{K_l}F_l^i, \mathbf{V_r}=W_d^{V_r}W_p^{V_r}F_r^i,	
	\end{split}
\end{equation}
\noindent where $\gamma_2$ is a trainable channel-wise scale parameter and initialized with zeros for stabilizing training, $W_p^r$ is the 1$\times$1 point-wise convolution, and $F_r^i$ denotes the right-view features extracted by the front BasicBlocks. After obtaining cross-view features $F_{l\leftarrow r}^i$ and $F_{r\leftarrow l}^i$, we concatenate them to $F_{l,r}^i$ and feed them into the next BasicBlocks.

\subsection{Loss Function}
We use two common loss functions to train our proposed StereoIRR model, i.e., perceptual loss and SSIM loss. The perceptual loss is defined as
\begin{equation} \label{eqn9}
	\mathcal{L}_{per}=\frac{1}{2}\sum_{i=1,2}\left \| \mathcal{F}_i\left (\hat{Y}_l, \hat{Y}_r\right )-\mathcal{F}_i\left ({Y}_l, {Y}_r\right )\right \|_2^2,
\end{equation}
\noindent where ${Y}_l$ and $\hat{Y}_l$ are the ground-truth and derained left-view images, ${Y}_r$ and $\hat{Y}_r$ are the ground-truth and derained right-view images, and $\mathcal{F}\left ( \cdot \right )$ is a non-linear CNN transformation. In our method, we use layer \textit{ReLU2\_2} and \textit{ReLU2\_3} of the VGG-16 model \cite{VGG} pretrained on ImageNet as $\mathcal{F}_1\left ( \cdot \right )$ and $\mathcal{F}_2\left ( \cdot \right )$. The SSIM loss is defined as follows: 
\begin{equation} \label{eqn10}
	\mathcal{L}_{ssim} = -SSIM\left((\hat{Y}_l, \hat{Y}_r), ({Y}_l, {Y}_r)\right),
\end{equation}
\noindent where $SSIM\left ( \cdot \right )$ calculates the similarity between images. 

\begin{table*}[ht]
	\centering
	\footnotesize
	\caption{\mdseries Ablation study on \textit{RainKITTI2012} dataset with different variants of StereoIRR. - indicates the setting is the same as the \textit{Baseline}.}
	\vspace{-2mm}%不可以换行
	\label{table:1}
	\linespread{1}\selectfont
	\begin{tabular}{c c c c c c c c c c}
		\toprule[1pt]
		\textbf{Models} & \textbf{\#Patch} & \textbf{\#Encoder} & \textbf{\#Middle} & \textbf{\#Decoder} & \textbf{\#Width} & \textbf{\#Loss} & \textbf{\#DMA} & \textbf{\#Scale} & \textbf{Total (PSNR/SSIM)} \\
		\midrule[0.5pt]
		\textit{Baseline} & 256 & [3, 3, 3, 3] & 1 & [3, 3, 3, 3] & 24 & Per.+SSIM & $\checkmark$ & $\checkmark$ & 38.939/0.9843 \\
		\textit{v1} & - & [4, 3, 3] & - & [3, 3, 4] & - & - & - & - & 38.346/0.9835 \\
		\textit{v2} & - & [3, 3, 3, 3, 3] & - & [3, 3, 3, 3, 3] & - & - & - & - & 39.105/0.9844 \\
		\textit{v3} & 336 & - & - & - & - & - & - & - & 39.329/0.9850 \\
		\textit{v4} & 320 & [3, 3, 3, 3, 3] & - & [3, 3, 3, 3, 3] & - & - & - & - & \underline{39.401/0.9854} \\
		\textit{v5} & - & - & - & - & 16 & - & - & - & 37.986/0.9817 \\
		\textit{v6} & - & - & - & - & 32 & - & - & - & 39.301/0.9852 \\
		\textit{v7} & 320 & [3, 3, 3, 3, 3] & - & [3, 3, 3, 3, 3] & 30 & - & - & - & \textbf{39.907/0.9864} \\
		\textit{v8} & - & - & 3 & - & - & - & - & - & 39.041/0.9845 \\
		\textit{v9} & 320 & [3, 3, 3, 3, 3] & - & [3, 3, 3, 3, 3] & 30 & MSE & - & - & 39.342/0.9844 \\
		\textit{v10} & 320 & [3, 3, 3, 3, 3] & - & [3, 3, 3, 3, 3] & 30 & - & \ding{55} & - & 38.153/0.9823 \\
		\textit{v11} & 320 & [3, 3, 3, 3, 3] & - & [3, 3, 3, 3, 3] & 30 & - & - & \ding{55} & 39.251/0.9845 \\
		\bottomrule[1pt]
	\end{tabular}
	\vspace{-1mm}%不可以换行
\end{table*}

\begin{table*}[t]
	\centering
	\footnotesize
	\vspace{2mm}%不可以换行
	\caption{\mdseries Evaluation (PSNR/SSIM) on RainKITTI2012 and RainKITTI2015 datasets. \textbf{Bold} and \underline{underline} denotes the best and the second.}
	\vspace{-2mm}%不可以换行
	\label{table:2}
	\linespread{1}\selectfont
	\begin{tabular}{c c c c c c c c c c}
		\toprule[1pt]
		~ & \multirow{3}{*}{{\textbf{Methods}}} & \multirow{3}{*}{\textbf{Venue}} & \multicolumn{3}{c}{{{\textbf{RainKITTI2012}}}} & ~ & \multicolumn{3}{c}{{{\textbf{RainKITTI2015}}}} \\
		\cmidrule[0.5pt]{4-6}
		\cmidrule[0.5pt]{8-10}
		~ & ~ & ~ & Left view & Right view & Total & ~ & Left view & Right view & Total \\
		\midrule[0.5pt]
		\multirow{5}{*}{\textbf{Monocular}} & DetailNet \cite{DetailNet} & cvpr'17 & 31.11/0.943 & 31.22/0.941 & 31.16/0.942 & ~ & 31.19/0.942 & 31.35/0.941 & 31.27/0.942 \\
		~ & RESCAN \cite{RESCAN} & eccv'18 & 34.48/0.970 & 34.14/0.969 & 34.31/0.970 & ~ & 34.43/0.968 & 34.19/0.968 & 34.31/0.968 \\
		~ & JORDER-E \cite{JORDER-E} & tpami'19 & 36.02/0.975 & 36.05/0.975 & 36.04/0.975 & ~ & 36.12/0.973 & 36.21/0.973 & 36.16/0.973 \\
		~ & RCDNet \cite{RCDNet} & cvpr'20 & 36.38/0.976 & 36.45/0.976 & 36.42/0.976 & ~ & 36.50/0.975 & 36.59/0.975 & 36.55/0.975 \\
		~ & MPRNet \cite{MPRNet} & cvpr'21 & 38.23/0.982 & 38.30/0.982 & 38.26/0.982 & ~ & 38.29/0.981 & 38.36/0.981 & 38.32/0.981 \\
		\midrule[0.5pt]		
		\multirow{4}{*}{\textbf{Stereo}} & iPASSR \cite{iPASSR} & cvpr'21 & 35.50/0.975 & 35.36/0.974 & 35.43/0.975 & ~ & 35.63/0.973 & 35.65/0.973 & 35.64/0.973 \\
		~ & NAFSSR \cite{NAFSSR} & cvpr'22 & \underline{38.64/0.983} & \underline{38.81/0.983} & \underline{38.73/0.983} & ~ & \underline{38.69/0.982} & \underline{38.92/0.983} & \underline{38.80/0.982} \\ 
		~ & EPRRNet \cite{EPRRNet} & ijcv'22 & 36.09/0.975 & 36.27/0.976 & 36.18/0.976 & ~ & 36.33/0.974 & 36.38/0.975 & 36.35/0.974 \\ 
		~ & StereoIRR (ours) & - & \textbf{39.80/0.986} & \textbf{40.01/0.987} & \textbf{39.91/0.986} & ~ & \textbf{39.49/0.983} & \textbf{39.75/0.985} & \textbf{39.62/0.984} \\
		\bottomrule[1pt]
	\end{tabular}
	\vspace{-2mm}%不可以换行
\end{table*}

\begin{table}[t]
	\centering
	\footnotesize
	\caption{\mdseries Evaluation (PSNR/SSIM) on StereoCityscapes dataset.}
	\vspace{-2mm}%不可以换行
	\label{table:3}
	\linespread{1}\selectfont
	\begin{tabular}{c c c c}
		\toprule[1pt]
		\multirow{3}{*}{{\textbf{Methods}}} & \multicolumn{3}{c}{{{\textbf{StereoCityscapes}}}}\\
		\cmidrule[0.5pt]{2-4}
		~ & Left view & Right view & Total \\
		\midrule[0.5pt]
		DetailNet \cite{DetailNet} & 27.79/0.910 & 27.43/0.909 & 27.61/0.910 \\
		RESCAN \cite{RESCAN} & 23.38/0.907 & 23.17/0.907 & 23.27/0.907 \\
		JORDER-E \cite{JORDER-E} & 30.80/0.964 & 29.76/0.961 & 30.28/0.963 \\
		RCDNet \cite{RCDNet} & 31.27/0.964 & 30.58/0.962 & 30.92/0.963 \\
		MPRNet \cite{MPRNet} & \underline{39.03/0.992} & \underline{37.82/0.991} & \underline{38.42/0.991} \\
		\midrule[0.5pt]
		iPASSR \cite{iPASSR} & 29.23/0.964 & 28.50/0.961 & 28.87/0.963 \\
		NAFSSR \cite{NAFSSR} & 36.34/0.988 & 35.62/0.987 & 35.98/0.988 \\ 
		EPRRNet \cite{EPRRNet} & 20.45/0.832 & 20.53/0.831 & 20.49/0.832 \\ 
		StereoIRR (ours) & \textbf{40.20/0.997} & \textbf{39.16/0.996} & \textbf{39.68/0.996} \\
		\bottomrule[1pt]
	\end{tabular}
	\vspace{-3mm}%不可以换行
\end{table}

\section{Experiments}
\subsection{Experimental Settings}
\textbf{Implementation details.}
The network is trained with PyTorch platform in the Python environment. All of the experiments including ablation studies are conducted on one NVIDIA GeForce GTX 3090 GPU with 24GB memory. Adam is used as the optimizer with $\beta_1$ = 0.9 and $\beta_2$ = 0.9 with weight decay 0 by default. The learning rate is set to 5e-4 and decreased every 50 epochs with the MultiStepLR strategy. The batch size is 3, and each image will be randomly cropped to 320$\times$320 pixels. We train 200 epochs to make the network convergence. 

\textbf{Compared methods.}
Five classic \textit{\textbf{monocular}} rain removal models are chosen for comparison, i.e., \textit{DetailNet} \cite{DetailNet}, \textit{RESCAN} \cite{RESCAN}, \textit{JORDER-E} \cite{JORDER-E}, \textit{RCDNet} \cite{RCDNet} and \textit{MPRNet} \cite{MPRNet}. Besides, three representative \textit{\textbf{stereo}} restoration methods, i.e., \textit{iPASSR} \cite{iPASSR}, \textit{NAFSSR} \cite{NAFSSR} and \textit{EPRRNet} \cite{EPRRNet} are also included for comparison. 

\textbf{Evaluation metrics.} Two classical Full-Reference Image Quality Assessments, i.e., PSNR \cite{PSNR} and SSIM \cite{SSIM}, are used. Note that we calculate the metrics on the Y channel with MatLab codes like the previous works \cite{JORDER-E, RCDNet}.

\textbf{Datasets}.
We use two popular synthetic stereo datasets, i.e., \textit{RainKITTI2012} and \textit{RainKITTI2015} \cite{PRRNet,EPRRNet} for evaluation. These two datasets are created using Photoshop based on the public KITTI dataset \cite{Kitti}. The training set and testing set of \textit{RainKITTI2012} and \textit{RainKITTI2015} contain 4,062/4,085 and 4,200/4,189 image pairs, respectively. We also use a physics-based rendering way \cite{Rendering2} to construct a new synthetic stereo dataset named \textit{StereoCityscapes} based on Cityscapes \cite{Cityscapes}. The training set and testing set of \textit{StereoCityscapes} contain 2975 and 1525 image pairs. 

%At present, \textit{SIRR-Data} \cite{SIRR} and \textit{SPA-Data} \cite{SPANet} are the two most commonly used real scenario datasets. We use them for generalization evaluation. 

\begin{figure*}[t]
	\includegraphics[width=\linewidth]{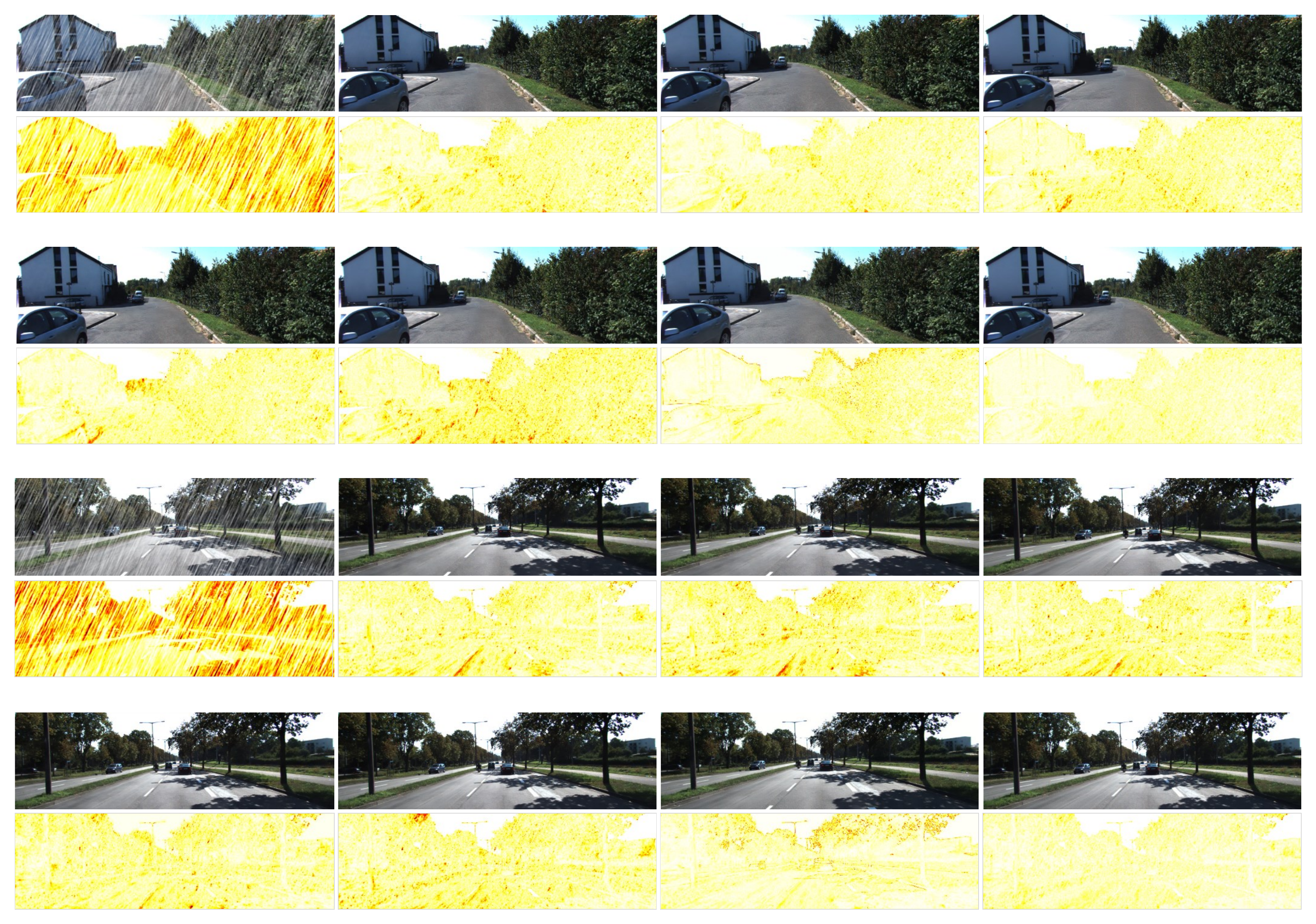}
	\begin{picture}(0,0)
		\put(19,272){\footnotesize Rain/error map (PSNR/SSIM)}
		\put(147,272){\footnotesize JORDER-E (36.55/0.977)}
		\put(273,272){\footnotesize RCDNet (36.60/0.977)}	
		\put(395,272){\footnotesize MPRNet (38.43/0.982)}
		\put(30,184){\footnotesize iPASSR (36.18/0.976)}
		\put(151,184){\footnotesize NAFSSR (38.61/0.983)}
		\put(271,184){\footnotesize EPRRNet (36.28/0.976)}
		\put(390,184){\footnotesize \textbf{StereoIRR (39.42/0.986)}}
		
		\put(19,96){\footnotesize Rain/error map (PSNR/SSIM)}
		\put(147,96){\footnotesize JORDER-E (36.06/0.974)}
		\put(273,96){\footnotesize RCDNet (36.52/0.975)}	
		\put(395,96){\footnotesize MPRNet (38.03/0.981)}
		\put(30,8){\footnotesize iPASSR (35.90/0.975)}
		\put(151,8){\footnotesize NAFSSR (38.65/0.983)}
		\put(271,8){\footnotesize EPRRNet (36.29/0.975)}
		\put(390,8){\footnotesize \textbf{StereoIRR (39.43/0.986)}}
	\end{picture}
	\vspace{-4mm}%不可以换行   
	\caption{Visualization of the derained images and the corresponding error maps of each method based on the RainKITTI2012 and RainKITTI2015 datasets, including JORDER-E \cite{JORDER-E}, RCDNet \cite{RCDNet}, MPRNet \cite{MPRNet}, iPASSR \cite{iPASSR}, NAFSSR \cite{NAFSSR}, EPRRNet \cite{EPRRNet} and our StereoIRR. Darker pixels in error maps indicate greater errors. Compared with other methods, our StereoIRR obtains the smallest errors. This demonstrates that StereoIRR can better remove the rain and restore the details.}\label{fig:6}
	\vspace{-3mm}%不可以换行
\end{figure*}
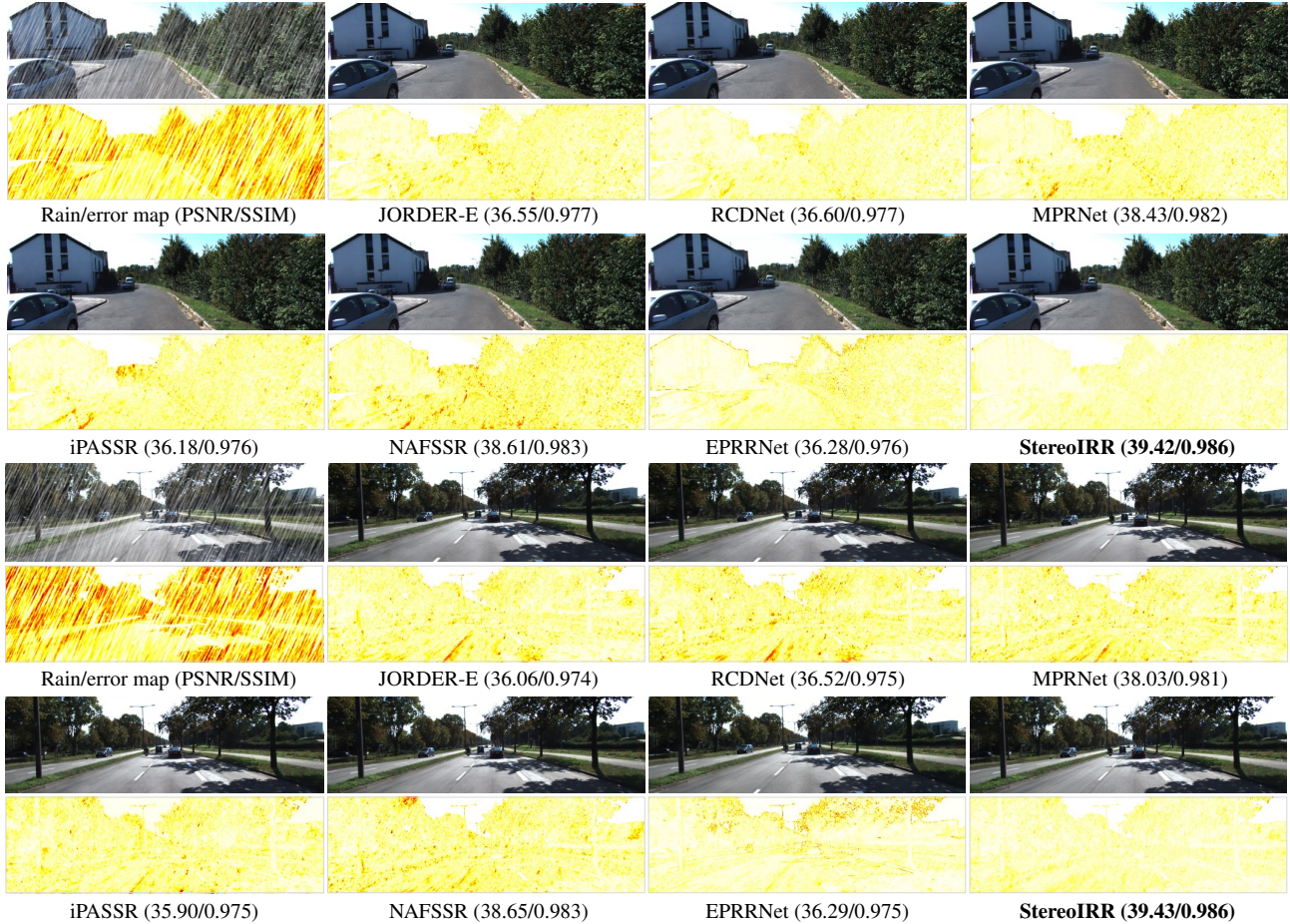

\subsection{Ablation Study}\label{Ablation}
We explore the impact of key factors in StereoIRR. We run all the ablation studies on a single GTX 3090 GPU with 24GB memory. We use variable control to test the performance on \textit{RainKITTI2012} dataset under each setting, and the results are described in Table \ref{table:1}. \textit{Baseline} represents the baseline model, from \textit{v1} to \textit{v11} are different models that change a single or several key factors, and - indicates that the key factor setting is as the same as the \textit{Baseline}. 

\textbf{Encoder-Decoder.}
StereoIRR uses a U-net architecture (Encoder-Middle-Decoder), where \textit{v1} and \textit{v2} are the shallower and deeper U-net architectures respectively. As can be seen, a deeper U-net architecture with long-range feature extraction can help the network learn more accurate disparity maps and thus achieve better performance. 

\textbf{Patch size.}
For the same Encoder-Decoder setting, increasing the patch size (\textit{v3}) greatly improves the performance. Because a larger patch size can save more feature information, which is conducive to subsequent up-sampling operations. Considering GPU memory limitation, we set Encoder-Decoder and patch size to [3, 3, 3, 3, 3] and 320, which obtains the second-best performance (\textit{v4}).

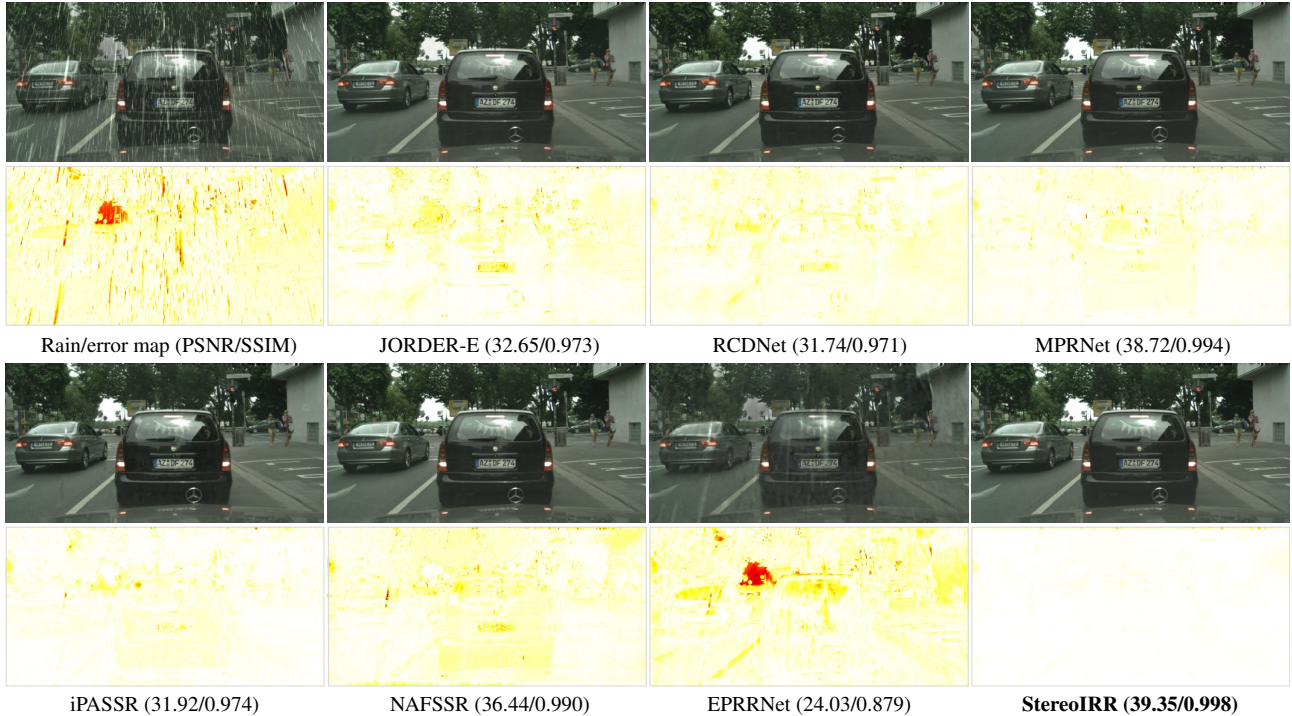
\begin{figure*}[ht!]
	\includegraphics[width=\linewidth]{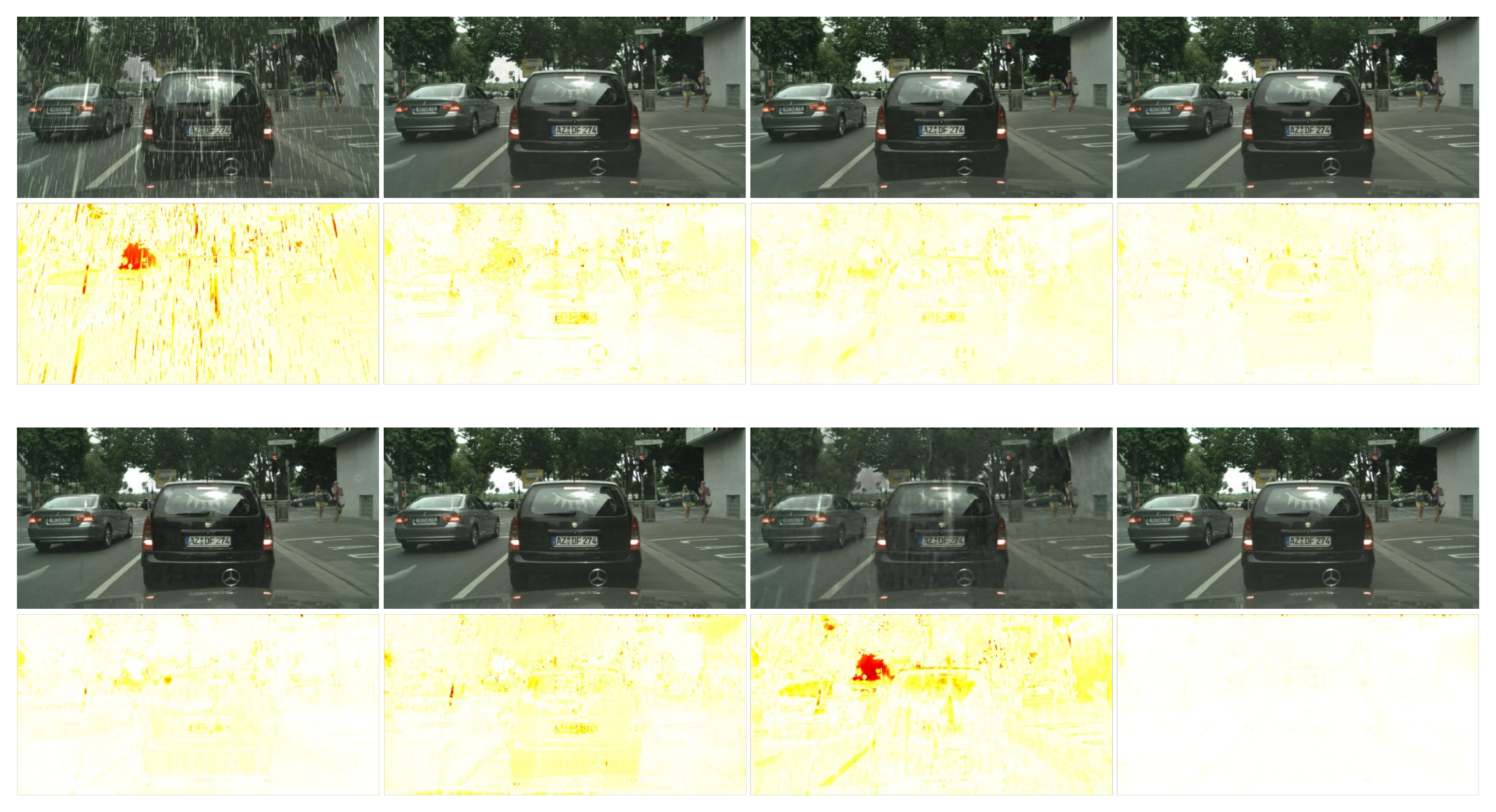}
	\begin{picture}(0,0)
		\put(19,144){\footnotesize Rain/error map (PSNR/SSIM)}
		\put(147,144){\footnotesize JORDER-E (32.65/0.973)}
		\put(273,144){\footnotesize RCDNet (31.74/0.971)}	
		\put(395,144){\footnotesize MPRNet (38.72/0.994)}
		\put(30,8){\footnotesize iPASSR (31.92/0.974)}
		\put(151,8){\footnotesize NAFSSR (36.44/0.990)}
		\put(271,8){\footnotesize EPRRNet (24.03/0.879)}
		\put(390,8){\footnotesize \textbf{StereoIRR (39.35/0.998)}}
	\end{picture}
	\vspace{-4mm}%不可以换行   
	\caption{Visualization of the derained images and the corresponding error maps of each method based on the StereoCityscapes dataset, including JORDER-E \cite{JORDER-E}, RCDNet \cite{RCDNet}, MPRNet \cite{MPRNet}, iPASSR \cite{iPASSR}, NAFSSR \cite{NAFSSR}, EPRRNet \cite{EPRRNet} and our StereoIRR. For the error maps, the whiter the better. We can indistinctly see a car in the error maps of all compared methods. In contrast, it is hard to see the shape of the car for our StereoIRR, i.e., less information is lost in our method.}
	\label{fig:7}
	\vspace{-3mm}%不可以换行
\end{figure*}

\textbf{Channel width.}
\textit{v5} and \textit{v6} are two kinds of channel width settings, and a larger channel width results in better performance than \textit{v1}. We set the channel width to 30 based on \textit{v4} and obtain the best result (\textit{v7}).

\textbf{Middle layer.}
We increased the numbers of middle layer in \textit{v8} and found that it gives a small boost compared with \textit{Baseline}. Due to the GPU memory limitations, we don't increase the numbers of middle layer in our StereoIRR.

\textbf{Loss function.}
Using MSE loss alone (\textit{v9}) results in a larger performance degradation compared to hybrid losses (\textit{v7}). Because perceptual loss can bring in more image details, while SSIM loss can ensure structural consistency.

\textbf{Dual-view mutual attention.}
We tested a variant without DMA (\textit{v10}) and found it obtains poor performance. This proves that a lack of cross-view feature fusion can not reasonably extract the complementary information between dual views, which is not conducive to rain removal. 

\textbf{Multi-scale.}
StereoIRR uses a U-net architecture with up/down-sampling. \textit{v11} shows a variant without up/down-sampling. The result shows that multiple scales can help the network to better mitigate the influence of rain streaks on complementary information in the long-range process.

\textbf{Remarks.}
Based on the ablation study, we decided on the network structure and training parameters of StereoIRR and applied them in the experimental comparison. 

%\subsection{Evaluation on Synthetic Datasets}
\subsection{Quantitative results}
We evaluate the performance of all the methods on three datasets, i.e., RainKITTI2012, RainKITTI2015, and StereoCityscapes. For a fair comparison, all the stereo data are entered as a single input for the monocular methods and as a pair input for the stereo methods. Table \ref{table:2} and Table \ref{table:3} show the numerical results of left view, right view and total. We see that our StereoIRR obtains better performance on all evaluated datasets. MPRNet \cite{MPRNet} and NAFSSR \cite{NAFSSR} obtain the best results in monocular and stereo compared methods, respectively. However, for MPRNet, it cannot use the complementary information between stereo views to recover image details, and for NAFSSR, its cross-view interaction mechanism is not as effective as our proposed DMA. In addition, the complex rain streak and rain fog in StereoCityscapes data made rain removal more difficult and resulted in greater performance differences among all the methods. Overall, the best PSNR/SSIM demonstrates that our StereoIRR is capable of restoring the details and recovering the structural information of stereo images.

\subsection{Visual analysis results}
For better comparison, we visualize the derained images with PSNR/SSIM matrices and corresponding error maps of each method on the RainKITTI2012 and RainKITTI2015 datasets in Figure \ref{fig:6}. Note that the process of computing the error maps can be referred to \cite{Error}. We see that our StereoIRR is capable of recovering more consistent structures and accurate textures than other competitors. Based on the error maps, StereoIRR is the lightest one, indicating that our model can recover more details. Figure \ref{fig:7} shows the visual comparison of the StereoCityscapes dataset. According to the displayed PSNR and SSIM, the proposed StereoIRR obtains maximum values, which suggests that the derained images of our model are of higher quality. Similar results can also be found in the error maps. The errors produced by our method are obviously smaller than others.

%\subsection{Evaluation on Real Scenario}
%To evaluate the deraining and segmentation capability of each SID method for real scenarios, we also conduct experiments on the proposed \textit{Cityscapes\_real} dataset. The experimental results are presented in Table \ref{table:3}. We can find that: \textbf{(1)} Our SGINet obtains nearly 1dB improvement on PSNR and 3dB improvement on MIoU, compared with RLNet. \textbf{(2)} Compared the \textit{Improvement} (\%) with RLNet over \textit{Cityscapes\_syn} dataset (deraining: \textit{3.6/0.4}, segementation: \textit{2.3/2.3}), our method also achieves a impressive \textit{Improvement} (\%) in \textit{Cityscapes\_real} dataset (deraining: \textit{2.1/0.8}, segementation: \textit{3.9/3.7}), which clearly proves that our method has stable performance and strong robustness in real scenario. We also illustrate some deraining and segmentation results in Fig. \ref{fig:6}. From the deraining perspective, our SGINet can remove more rain streaks and recover the detailed structure and content of rain image. From the segmentation perspective, our SGINet delivers better semantic information than other methods, e.g., the person walking in the middle of image. 

\section{Conclusion and Future Work}
We have discussed the issue of discovering complementary information and estimating the disparity between stereo images by a new perspective of dual-view mutual attention. Technically, we proposed a new stereo image rain removal method named StereoIRR, which effectively addresses the stereo rain removal task with complex distributions via a long-range and cross-view interaction process, to alleviate the adverse effect of rain on complementary information and enable the stereo image features to get long-range and cross-view fusion. We present a new dual-view mutual attention mechanism that can generate mutual attention maps by facilitating cross-view feature fusion by taking left and right views as key information for each other. StereoIRR achieves superior performance to both monocular and stereo image rain removal methods on several widely-used datasets. In the future, we will extend the stereo image deraining task to joint high-level task.

\section{Acknowledgments}
This work is partially supported by the National Natural Science Foundation of China (62072151, 61732007, 61932009, 2020106007), and the Anhui Provincial Natural Science Fund for the Distinguished Young Scholars (2008085J30). Corresponding author: Zhao Zhang. 

%%%%%%%%% REFERENCES
{\small
	\bibliographystyle{ieee_fullname}
	\bibliography{egbib}

\begin{thebibliography}{10}\itemsep=-1pt

\bibitem{LN}
Jimmy~Lei Ba, Jamie~Ryan Kiros, and Geoffrey~E Hinton.
\newblock Layer normalization.
\newblock {\em arXiv preprint arXiv:1607.06450}, 2016.

\bibitem{SSIM}
Alan~C Brooks, Xiaonan Zhao, and Thrasyvoulos~N Pappas.
\newblock Structural similarity quality metrics in a coding context: exploring
  the space of realistic distortions.
\newblock {\em IEEE Transactions on image processing}, 17(8):1261--1273, 2008.

\bibitem{NAFNet}
Liangyu Chen, Xiaojie Chu, Xiangyu Zhang, and Jian Sun.
\newblock Simple baselines for image restoration.
\newblock {\em arXiv preprint arXiv:2204.04676}, 2022.

\bibitem{NAFSSR}
Xiaojie Chu, Liangyu Chen, and Wenqing Yu.
\newblock Nafssr: Stereo image super-resolution using nafnet.
\newblock In {\em Proceedings of the IEEE/CVF Conference on Computer Vision and
  Pattern Recognition}, pages 1239--1248, 2022.

\bibitem{Cityscapes}
Marius Cordts, Mohamed Omran, Sebastian Ramos, Timo Rehfeld, Markus Enzweiler,
  Rodrigo Benenson, Uwe Franke, Stefan Roth, and Bernt Schiele.
\newblock The cityscapes dataset for semantic urban scene understanding.
\newblock In {\em Proceedings of the IEEE Conference on Computer Vision and
  Pattern Recognition}, pages 3213--3223, 2016.

\bibitem{Stereo_SR6}
Qinyan Dai, Juncheng Li, Qiaosi Yi, Faming Fang, and Guixu Zhang.
\newblock Feedback network for mutually boosted stereo image super-resolution
  and disparity estimation.
\newblock In {\em Proceedings of the 29th ACM International Conference on
  Multimedia}, pages 1985--1993, 2021.

\bibitem{DetailNet}
Xueyang Fu, Jiabin Huang, Delu Zeng, Yue Huang, Xinghao Ding, and John Paisley.
\newblock Removing rain from single images via a deep detail network.
\newblock In {\em Proceedings of the IEEE Conference on Computer Vision and
  Pattern Recognition}, pages 3855--3863, 2017.

\bibitem{Kitti}
Andreas Geiger, Philip Lenz, Christoph Stiller, and Raquel Urtasun.
\newblock Vision meets robotics: The kitti dataset.
\newblock {\em The International Journal of Robotics Research},
  32(11):1231--1237, 2013.

\bibitem{GAN}
Ian Goodfellow, Jean Pouget-Abadie, Mehdi Mirza, Bing Xu, David Warde-Farley,
  Sherjil Ozair, Aaron Courville, and Yoshua Bengio.
\newblock Generative adversarial nets.
\newblock {\em Advances in neural information processing systems}, 27, 2014.

\bibitem{Rendering2}
Shirsendu~Sukanta Halder, Jean-Fran{\c{c}}ois Lalonde, and Raoul~de Charette.
\newblock Physics-based rendering for improving robustness to rain.
\newblock In {\em Proceedings of the IEEE International Conference on Computer
  Vision}, pages 10203--10212, 2019.

\bibitem{GELU}
Dan Hendrycks and Kevin Gimpel.
\newblock Gaussian error linear units (gelus).
\newblock {\em arXiv preprint arXiv:1606.08415}, 2016.

\bibitem{LSTM}
Sepp Hochreiter and J{\"u}rgen Schmidhuber.
\newblock Long short-term memory.
\newblock {\em Neural computation}, 9(8):1735--1780, 1997.

\bibitem{CA}
Jie Hu, Li Shen, and Gang Sun.
\newblock Squeeze-and-excitation networks.
\newblock In {\em Proceedings of the IEEE conference on computer vision and
  pattern recognition}, pages 7132--7141, 2018.

\bibitem{DAF-Net}
Xiaowei Hu, Chi-Wing Fu, Lei Zhu, and Pheng-Ann Heng.
\newblock Depth-attentional features for single-image rain removal.
\newblock In {\em Proceedings of the IEEE Conference on Computer Vision and
  Pattern Recognition}, pages 8022--8031, 2019.

\bibitem{PSNR}
Quan Huynh-Thu and Mohammed Ghanbari.
\newblock Scope of validity of psnr in image/video quality assessment.
\newblock {\em Electronics letters}, 44(13):800--801, 2008.

\bibitem{Stereo_SR1}
Daniel~S Jeon, Seung-Hwan Baek, Inchang Choi, and Min~H Kim.
\newblock Enhancing the spatial resolution of stereo images using a parallax
  prior.
\newblock In {\em Proceedings of the IEEE conference on computer vision and
  pattern recognition}, pages 1721--1730, 2018.

\bibitem{MSPFN}
Kui Jiang, Zhongyuan Wang, Peng Yi, Chen Chen, Baojin Huang, Yimin Luo, Jiayi
  Ma, and Junjun Jiang.
\newblock Multi-scale progressive fusion network for single image deraining.
\newblock In {\em Proceedings of the IEEE/CVF Conference on Computer Vision and
  Pattern Recognition}, pages 8346--8355, 2020.

\bibitem{DeHRain}
Ruoteng Li, Loong-Fah Cheong, and Robby~T Tan.
\newblock Heavy rain image restoration: Integrating physics model and
  conditional adversarial learning.
\newblock In {\em Proceedings of the IEEE Conference on Computer Vision and
  Pattern Recognition}, pages 1633--1642, 2019.

\bibitem{RESCAN}
Xia Li, Jianlong Wu, Zhouchen Lin, Hong Liu, and Hongbin Zha.
\newblock Recurrent squeeze-and-excitation context aggregation net for single
  image deraining.
\newblock In {\em Proceedings of the European Conference on Computer Vision
  (ECCV)}, pages 254--269, 2018.

\bibitem{UBCN}
Yi Li, Yi Chang, Changfeng Yu, and Luxin Yan.
\newblock Close the loop: a unified bottom-up and top-down paradigm for joint
  image deraining and segmentation.
\newblock In {\em Proceedings of the AAAI Conference on Artificial
  Intelligence}, pages 1--9, 2022.

\bibitem{GMM}
Yu Li, Robby~T Tan, Xiaojie Guo, Jiangbo Lu, and Michael~S Brown.
\newblock Rain streak removal using layer priors.
\newblock In {\em Proceedings of the IEEE conference on computer vision and
  pattern recognition}, pages 2736--2744, 2016.

\bibitem{SwinIR}
Jingyun Liang, Jiezhang Cao, Guolei Sun, Kai Zhang, Luc Van~Gool, and Radu
  Timofte.
\newblock Swinir: Image restoration using swin transformer.
\newblock In {\em Proceedings of the IEEE/CVF International Conference on
  Computer Vision}, pages 1833--1844, 2021.

\bibitem{DSC}
Yu Luo, Yong Xu, and Hui Ji.
\newblock Removing rain from a single image via discriminative sparse coding.
\newblock In {\em Proceedings of the IEEE International Conference on Computer
  Vision}, pages 3397--3405, 2015.

\bibitem{RICNet}
Siqi Ni, Xueyun Cao, Tao Yue, and Xuemei Hu.
\newblock Controlling the rain: From removal to rendering.
\newblock In {\em Proceedings of the IEEE/CVF Conference on Computer Vision and
  Pattern Recognition}, pages 6328--6337, 2021.

\bibitem{Stereo_DH2}
Jing Nie, Yanwei Pang, Jin Xie, Jing Pan, and Jungong Han.
\newblock Stereo refinement dehazing network.
\newblock {\em IEEE Transactions on Circuits and Systems for Video Technology},
  32(6):3334--3345, 2021.

\bibitem{Stereo_DH1}
Yanwei Pang, Jing Nie, Jin Xie, Jungong Han, and Xuelong Li.
\newblock Bidnet: Binocular image dehazing without explicit disparity
  estimation.
\newblock In {\em Proceedings of the IEEE/CVF conference on computer vision and
  pattern recognition}, pages 5931--5940, 2020.

\bibitem{CCN}
Ruijie Quan, Xin Yu, Yuanzhi Liang, and Yi Yang.
\newblock Removing raindrops and rain streaks in one go.
\newblock In {\em Proceedings of the IEEE/CVF Conference on Computer Vision and
  Pattern Recognition}, pages 1--10, 2021.

\bibitem{U-net}
Olaf Ronneberger, Philipp Fischer, and Thomas Brox.
\newblock U-net: Convolutional networks for biomedical image segmentation.
\newblock In {\em International Conference on Medical image computing and
  computer-assisted intervention}, pages 234--241. Springer, 2015.

\bibitem{Stereo_DR5}
Zifan Shi, Na Fan, Dit-Yan Yeung, and Qifeng Chen.
\newblock Stereo waterdrop removal with row-wise dilated attention.
\newblock In {\em 2021 IEEE/RSJ International Conference on Intelligent Robots
  and Systems (IROS)}, pages 3829--3836. IEEE, 2021.

\bibitem{VGG}
Karen Simonyan and Andrew Zisserman.
\newblock Very deep convolutional networks for large-scale image recognition.
\newblock {\em arXiv preprint arXiv:1409.1556}, 2014.

\bibitem{Attention}
Ashish Vaswani, Noam Shazeer, Niki Parmar, Jakob Uszkoreit, Llion Jones,
  Aidan~N Gomez, {\L}ukasz Kaiser, and Illia Polosukhin.
\newblock Attention is all you need.
\newblock In {\em Advances in neural information processing systems}, pages
  5998--6008, 2017.

\bibitem{RCDNet}
Hong Wang, Qi Xie, Qian Zhao, and Deyu Meng.
\newblock A model-driven deep neural network for single image rain removal.
\newblock In {\em Proceedings of the IEEE/CVF Conference on Computer Vision and
  Pattern Recognition}, pages 3103--3112, 2020.

\bibitem{SPANet}
Tianyu Wang, Xin Yang, Ke Xu, Shaozhe Chen, Qiang Zhang, and Rynson~WH Lau.
\newblock Spatial attentive single-image deraining with a high quality real
  rain dataset.
\newblock In {\em Proceedings of the IEEE Conference on Computer Vision and
  Pattern Recognition}, pages 12270--12279, 2019.

\bibitem{iPASSR}
Yingqian Wang, Xinyi Ying, Longguang Wang, Jungang Yang, Wei An, and Yulan Guo.
\newblock Symmetric parallax attention for stereo image super-resolution.
\newblock In {\em Proceedings of the IEEE/CVF Conference on Computer Vision and
  Pattern Recognition}, pages 766--775, 2021.

\bibitem{Uformer}
Zhendong Wang, Xiaodong Cun, Jianmin Bao, Wengang Zhou, Jianzhuang Liu, and
  Houqiang Li.
\newblock Uformer: A general u-shaped transformer for image restoration.
\newblock In {\em Proceedings of the IEEE/CVF Conference on Computer Vision and
  Pattern Recognition}, pages 17683--17693, 2022.

\bibitem{SIRR}
Wei Wei, Deyu Meng, Qian Zhao, Zongben Xu, and Ying Wu.
\newblock Semi-supervised transfer learning for image rain removal.
\newblock In {\em Proceedings of the IEEE Conference on Computer Vision and
  Pattern Recognition}, pages 3877--3886, 2019.

\bibitem{DerainCycleGAN}
Yanyan Wei, Zhao Zhang, Yang Wang, Mingliang Xu, Yi Yang, Shuicheng Yan, and
  Meng Wang.
\newblock Deraincyclegan: Rain attentive cyclegan for single image deraining
  and rainmaking.
\newblock {\em IEEE Transactions on Image Processing}, 30:4788--4801, 2021.

\bibitem{RadNet}
Yanyan Wei, Zhao Zhang, Mingliang Xu, Richang Hong, Jicong Fan, and Shuicheng
  Yan.
\newblock Robust attention deraining network for synchronous rain streaks and
  raindrops removal.
\newblock In {\em Proceedings of the 30th ACM International Conference on
  Multimedia}, pages 6464--6472, 2022.

\bibitem{SGINet}
Yanyan Wei, Zhao Zhang, Huan Zheng, Richang Hong, Yi Yang, and Meng Wang.
\newblock Sginet: Toward sufficient interaction between single image deraining
  and semantic segmentation.
\newblock In {\em Proceedings of the 30th ACM International Conference on
  Multimedia}, pages 6202--6210, 2022.

\bibitem{Stereo_SR4}
Bo Yan, Chenxi Ma, Bahetiyaer Bare, Weimin Tan, and Steven~CH Hoi.
\newblock Disparity-aware domain adaptation in stereo image restoration.
\newblock In {\em Proceedings of the IEEE/CVF Conference on Computer Vision and
  Pattern Recognition}, pages 13179--13187, 2020.

\bibitem{JORDER-E}
Wenhan Yang, Robby~T Tan, Jiashi Feng, Jiaying Liu, Shuicheng Yan, and Zongming
  Guo.
\newblock Joint rain detection and removal from a single image with
  contextualized deep networks.
\newblock {\em IEEE transactions on pattern analysis and machine intelligence},
  2019.

\bibitem{Restormer}
Syed~Waqas Zamir, Aditya Arora, Salman Khan, Munawar Hayat, Fahad~Shahbaz Khan,
  and Ming-Hsuan Yang.
\newblock Restormer: Efficient transformer for high-resolution image
  restoration.
\newblock In {\em Proceedings of the IEEE/CVF Conference on Computer Vision and
  Pattern Recognition}, pages 5728--5739, 2022.

\bibitem{MPRNet}
Syed~Waqas Zamir, Aditya Arora, Salman Khan, Munawar Hayat, Fahad~Shahbaz Khan,
  Ming-Hsuan Yang, and Ling Shao.
\newblock Multi-stage progressive image restoration.
\newblock In {\em Proceedings of the IEEE conference on computer vision and
  pattern recognition}, 2021.

\bibitem{DID-MDN}
He Zhang and Vishal~M Patel.
\newblock Density-aware single image de-raining using a multi-stream dense
  network.
\newblock In {\em Proceedings of the IEEE conference on computer vision and
  pattern recognition}, pages 695--704, 2018.

\bibitem{PRRNet}
Kaihao Zhang, Wenhan Luo, Wenqi Ren, Jingwen Wang, Fang Zhao, Lin Ma, and
  Hongdong Li.
\newblock Beyond monocular deraining: Stereo image deraining via semantic
  understanding.
\newblock In {\em European Conference on Computer Vision}, pages 71--89.
  Springer, 2020.

\bibitem{EPRRNet}
Kaihao Zhang, Wenhan Luo, Yanjiang Yu, Wenqi Ren, Fang Zhao, Changsheng Li, Lin
  Ma, Wei Liu, and Hongdong Li.
\newblock Beyond monocular deraining: Parallel stereo deraining network via
  semantic prior.
\newblock {\em International Journal of Computer Vision}, pages 1--16, 2022.

\bibitem{Error}
Chuanjun Zheng, Daming Shi, and Yukun Liu.
\newblock Windowing decomposition convolutional neural network for image
  enhancement.
\newblock In {\em Proceedings of the 29th ACM International Conference on
  Multimedia}, pages 424--432, 2021.

\bibitem{Stereo_DB1}
Shangchen Zhou, Jiawei Zhang, Wangmeng Zuo, Haozhe Xie, Jinshan Pan, and
  Jimmy~S Ren.
\newblock Davanet: Stereo deblurring with view aggregation.
\newblock In {\em Proceedings of the IEEE/CVF Conference on Computer Vision and
  Pattern Recognition}, pages 10996--11005, 2019.

\end{thebibliography}
}

\end{document}